\documentclass{article}
\usepackage{spconf,amsmath,graphicx}
\usepackage{wrapfig}
\usepackage{times}
\usepackage{epsfig}
\usepackage{graphicx}
\usepackage{amsmath}
\usepackage{amssymb}

\usepackage{multirow}
\usepackage{makecell}

\usepackage{caption}
\usepackage{subcaption}
\usepackage{algorithm2e}
\usepackage{algpseudocode}
\usepackage{float}
\usepackage{url}

\usepackage{color}
\usepackage{xcolor,colortbl}
\usepackage{framed}
\usepackage{enumitem}
\usepackage{booktabs}
\usepackage[pagebackref=true,breaklinks=true,letterpaper=true,colorlinks,bookmarks=false]{hyperref}

\title{Solving Missing-Annotation Object Detection with\\ Background Recalibration Loss}
\name{Han Zhang$^{1}$ \ \ \ \ Fangyi Chen$^{1}$ \ \ \ \ Zhiqiang Shen$^{1\dagger}$\thanks{$^{\dagger}$Corresponding author.} \ \ \ \ Qiqi Hao$^{2}$ \ \ \ \ Chenchen Zhu$^{1}$\ \ \ \  Marios Savvides$^{1}$}

\address{$^{1}$Carnegie Mellon University\\ $^{2}$Beihang University}
\begin{document}
%\ninept
%
\maketitle
\begin{abstract}
This paper focuses on a novel and challenging detection scenario: A majority of true objects/instances is unlabeled in the datasets, so these missing-labeled areas will be regarded as the background during training. Previous art~\cite{wu2018soft} on this problem has proposed to use soft sampling to re-weight the gradients of RoIs based on the overlaps with positive instances, while their method is mainly based on the two-stage detector (i.e. Faster RCNN) which is more robust and friendly for the missing label scenario. In this paper, we introduce a superior solution called {\em Background Recalibration Loss} ({BRL}) that can automatically re-calibrate the loss signals according to the pre-defined IoU threshold and input image. Our design is built on the one-stage detector which is faster and lighter. Inspired by the {\em Focal Loss}~\cite{lin2017focal} formulation, we make several significant modifications to fit on the missing-annotation circumstance. We conduct extensive experiments on the curated PASCAL VOC~\cite{everingham2015pascal} and MS COCO~\cite{lin2014microsoft} datasets. The results demonstrate that our proposed method outperforms the baseline and other state-of-the-arts by a large margin. Code available: \url{https://github.com/Dwrety/mmdetection-selective-iou}.

%The proposed BRL differs from existing systems in three technical aspects: (1) 
\end{abstract}
\section{Introduction}
\label{sec:intro}
In real cases, generic object detection always faces the challenge of annotation quality, meaning that only partial instances are well annotated and a large proportion of true objects are missed, especially when the size of collected datasets become larger and larger. A good example of such a problem is OpenImages V4~\cite{kuznetsova2018open}, which contains 9.2M images, 15.4M bounding boxes across 600 object classes (more than 10$\times$ larger than COCO dataset~\cite{lin2014microsoft}).
On such a large dataset, it's fairly impossible to annotate every existing object in each image in practice. Thus, how to develop an automatic method to tackle this issue has been more and more important and also an urgent and promising direction to break through.

\begin{figure}[htb]
\centering
    \begin{minipage}[b]{1.0\linewidth}
    \centering
    \centerline{\includegraphics[width=8.cm]{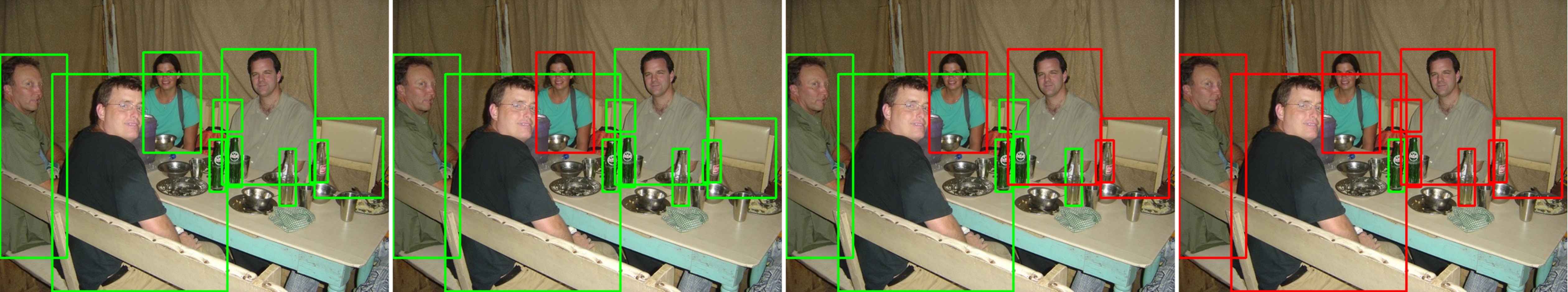}}
    % \vspace{0.5cm}
    \end{minipage}

    \begin{minipage}[b]{1.0\linewidth}
    \centering
    \centerline{\includegraphics[width=8.cm]{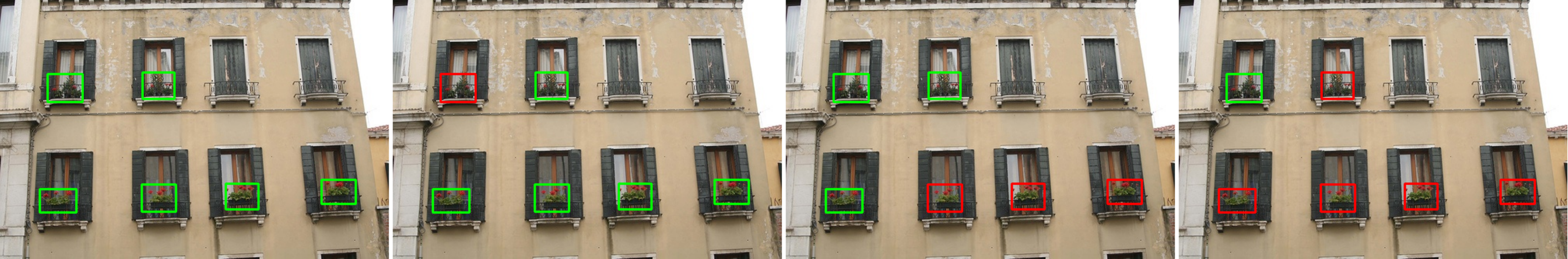}}
    % \vspace{0.5cm}
    \end{minipage}
    
    \begin{minipage}[b]{1.0\linewidth}
    \centering
    \centerline{\includegraphics[width=8.cm]{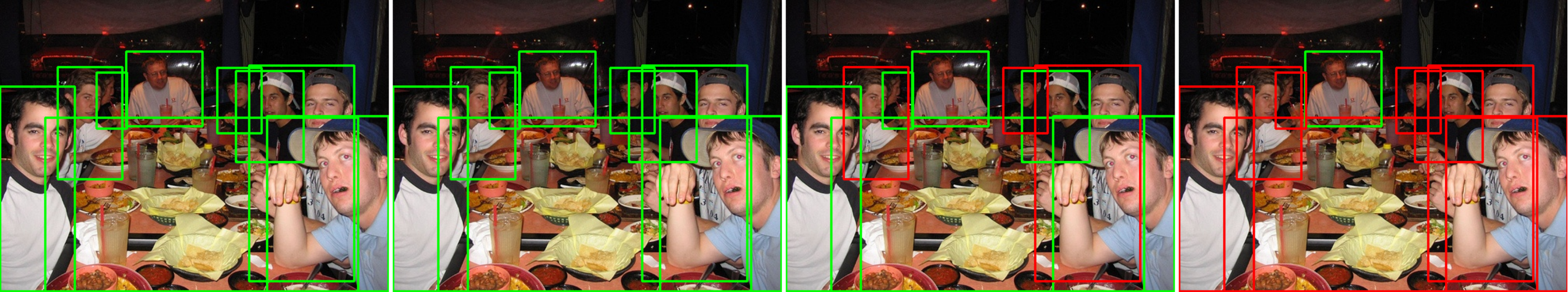}}
    % \vspace{0.5cm}
    \end{minipage}
    
    \begin{minipage}[b]{1.0\linewidth}
    \centering
    \centerline{\includegraphics[width=8.cm]{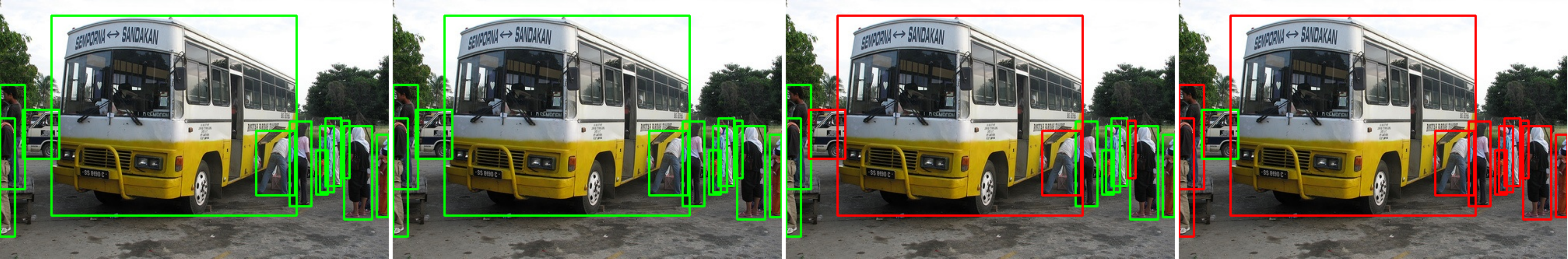}}
    % \vspace{0.1cm}
    \end{minipage}
\vspace{-0.5cm}
\caption{Dataset with missing labels. In this paper, we modified the PASCAL VOC dataset. Each column represents from left to right the \textit{\textbf{normal}}, \textit{\textbf{easy}}, \textit{\textbf{hard}}, and \textit{\textbf{extreme}} settings in Sec.~\ref{ssec:baseline}. The green bounding boxes are the ground truth annotations in the original dataset, while the red bounding boxes are erased during training. Zoom in for better resolution.}
\vspace{-0.15in}
\label{fig:dataset}
\end{figure}

The major challenge with the missing-annotated dataset is that a detector would suffer from incorrect supervision signals due to objects missing annotations, i.e., the unlabeled object areas will be treated as background and confuse the model during training. To further take the incompleteness of annotations into account, one straight-forward way is to use other trained detector from well-annotated dataset (such as COCO~\cite{lin2014microsoft}) to detect the presence of the amphibolous objects. The fatal drawback in this solution is that the label distributions (categories) in diverse datasets are usually different (like PASCAL VOC~\cite{everingham2015pascal} has 20, COCO~\cite{lin2014microsoft} has 80 and OpenImage V4~\cite{kuznetsova2018open} has 600 classes). It's hard or even unpractical to directly transfer detectors from one dataset to another. Recent advances in this field like {\em part-aware} sampling~\cite{niitani2019sampling}, {\em soft} sampling~\cite{wu2018soft}, etc. have proposed to use re-sampling based method to handle this problem, which is more efficient and practical. Specifically, {\em part-aware} sampling~\cite{niitani2019sampling} proposed the method that used human intuition for the hierarchical relation between labeled and unlabeled objects. {\em soft sampling} ~\cite{wu2018soft} re-weighted the gradients of RoIs as a pre-defined function of overlap with positive objects to make sure that the uncertain background areas are assigned a smaller training weight than the hard-negatives.

In this paper, we propose a more elegant solution called {\em Background Recalibration Loss} (BRL) that can automatically re-calibrate the training signals according to the pre-defined IoU thresholds and the content of input image. Our proposed method is straight-forward yet effective, easy to implement and can help improve the baseline vastly.
To demonstrate the effectiveness of the proposed {\em Background Recalibration Loss}, we implement function in the well-known RetinaNet~\cite{lin2017focal}, as we will introduce later, one-stage detectors are intrinsically less robust to missing labels. Extensive results on PASCAL VOC and COCO show that our proposed method outperforms the baseline and other state-of-the-arts by a large margin.

\section{Related Work}
\label{sec:related}
% \noindent{\textbf{Generic Object Detection.}}
In the generic object detection, proposal based (i.e. two-stage) methods include R-CNN~\cite{girshick14CVPR}, Fast RCNN~\cite{girshickICCV15fastrcnn}, Faster R-CNN~\cite{ren2015faster}, Mask R-CNN~\cite{he2017mask}, etc., are the most popular detectors in many real-world applications, such as surveillance, self-driving, etc. Another branch for general object detection is the one-stage detectors, such as YOLO~\cite{Redmon_2016}, SSD~\cite{liu2016ssd}, RetinaNet~\cite{lin2017focal}, DSOD~\cite{Shen2017DSOD}, FSAF~\cite{Zhu2019FeatureSA}, etc. In this paper, we adopt one-stage detector and propose a novel loss design for the missing-label circumstance.

Re-sampling with imbalanced or unlabeled objects is a common approach for object detection in the recent years. The representative methods among them like GHMC~\cite{li2019gradient}, OHEM~\cite{shrivastavaCVPR16ohem} and Focal Loss~\cite{lin2017focal} have been proposed for sampling RoIs and re-weighting the gradients. For instance, Li et al.~\cite{li2019gradient} proposed the gradient harmonizing mechanism (GHM) method that can down-weight the easy samples and the outliers for better model convergence. Shrivastava et al.~\cite{shrivastavaCVPR16ohem} presented the online hard example mining (OHEM) method based on the motivation that detection datasets always contain a large number of easy samples and a limited number of hard samples, training with more hard samples can help to improve the efficiency of learning. Our proposed method is motivated by Focal Loss~\cite{lin2017focal}, while we present a new formation with {\em Background Recalibration} to make it more effective and reasonable for the partial annotation scenarios.

\section{Rethinking Missing-label Object Detection}
\noindent{\textbf{Object Detection in Missing Annotation Scenario.}} Training object detectors with incompleteness of annotations is a new rising and challenging vision problem that aims to learn a robust detector by recalibrating the incorrect training signal with partial annotations. In the recent years, a variety of two-stage detector based methods have been proposed, such as {\em part-aware} sampling~\cite{niitani2019sampling} and {\em soft} sampling~\cite{wu2018soft}. We observe that two-stage is naturally more robust than the one-stage detectors for the missing-annotation circumstance. This advantage originates from the training strategy. One common practice for training two-stage detectors is to randomly sample a balanced batch of positive and negative examples during each iteration. Due to the great population of negative anchors, the model can hardly be affected by the limited propagation of errors in most sampling cases. That's why most of the previous solutions are built on the two-stage detectors. 

However, two-stage detectors have some intrinsic drawbacks such as 1) usually slower than one-stage detectors since two-stage detectors require large-resolution input size to maintain the high performance; 2) too complicated to adjust the hyper-parameters and not flexible to different datasets; 3) in some extreme cases of missing labels, the benefits from random sampling will still reach its limits. The question is what if we can make single-stage detectors more robust such that they can inherit the faster speed and accuracy advantages and be more tolerant to difficult data. Unlike two-stage detectors, the common practice for training anchor-based single-stage detectors is to use either hard example mining or to not use sampling at all. Recent success in RetinaNet \cite{lin2017focal} shows the model can be trained with all the anchors at once with huge class imbalance and still achieves high performance. Nevertheless, the problem becomes difficult when the majority of the annotations are missing. To make this idea applicable, we develop the approaches to filter out these error signals and correct them if possible.

\vspace{-0.05in}

% \noindent{\textbf{The difference with semi-supervised learning.}}
% \\

\section{Proposed Method}
\label{sec:method}
Single stage detectors often depend on a dense coverage of anchors on the image feature maps. While the dense coverage ensures sufficient training samples, it naturally introduces massive class imbalance. {Focal Loss} is widely favored among single stage detectors as it effectively re-scales the gradients of the anchors. In general, it calibrates and re-scales the gradients for both hard and easy training examples. However, any outlier will also spike the gradients as they are usually hard examples. In the case of missing labels, we hypothesize that these missing annotations are a type of hard negative examples because their feature-level resemblance to the positive ones. 

\begin{figure}[htb]
\centering
    \begin{minipage}[b]{1.0\linewidth}
    \centerline{\includegraphics[width=6.cm]{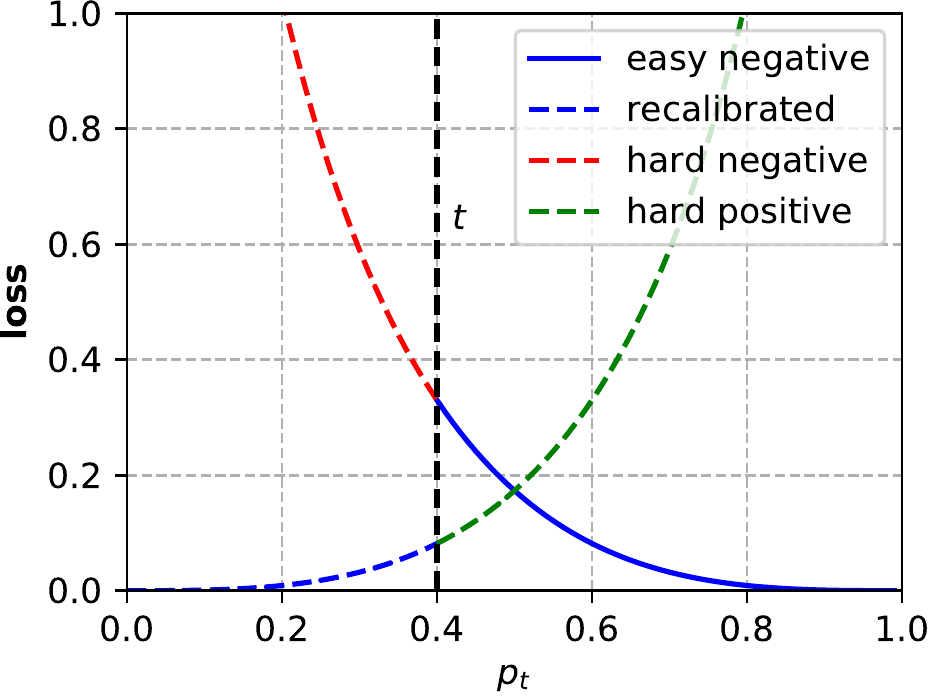}}
    \vspace{-0.3cm}
    \end{minipage}
    \caption{The plot of Background Recalibration Loss. The negative branch of the {\em focal loss} is changed to the mirrored positive branch below a confusion threshold $t$. During training, we apply the loss to the confusing anchors.}
    \label{illustration}
\vspace{-0.5cm}
\end{figure}

One quick solution is to perform IoU threshold by taking into account only the anchors with sufficient IoU values with the existing ground truth bounding boxes. Any other anchors are marked as confusion anchors and ignored during training. This raises the problem that the background information is lost during this brute-force mining. Despite the fact that we can extract most of these confusion anchors by IoU threshold, it is difficult to isolate them from the large pool of true-negative backgrounds. Therefore, we propose the {\em Background Recalibration Loss} which adjusts the gradient direction according to its own activation to reduce the adverse effect of error signals. But first, we briefly review the {Focal Loss}: 
\begin{equation}
{FL}(p_{t}) = -\alpha_{t} (1 - p_{t})^{\gamma}log(p_{t})
\end{equation} The terms $\alpha_{t}$ and $\gamma$ are the scaling factors for loss and gradients. The term ${p_{t}} \in [0,1]$, represents the predicted confidence score of a anchor. For our task, we are mainly interested in the branch for the negative samples. The meaning of ${p_{t}}$ is straightforward. The greater the value, higher the probability that the anchor is predicted as negative. On the contrary, the anchors associated with missing annotations would have lower activation as negative and generate huge gradients if ${p_{t}}$ is close to $0$. Directly ignoring these hard negative samples appears to be a good solution, but we take a step further by proposing a better gradient control method through the recalibration of the hard negative samples. We change the negative branch of the \textit{focal loss} by replacing it with the mirrored positive branch when the activation is below some confusion threshold $t$, as illustrated in Fig.~\ref{illustration}:  
\begin{equation}
    BRL(p_{t}) = 
    \begin{cases}
        -\alpha_{t} (1 - p_{t})^{\gamma}log(p_{t}), & p_{t} > t \\ \\
        -\alpha_{t} p_{t}^{\gamma}log(1-p_{t}), & {otherwise}
    \end{cases}
\end{equation}

The motivation of using the mirrored positive branch is directly related to our task. Generally, hard negative examples are the exact counterparts of easy positive examples in term of their feature level representations. Despite the lack of correct labels, the model can adjust the gradients according to its own well established classifier. Specifically, if the feature map of an anchor region is similar to that of a ground truth object, the classifier naturally assigns a low confidence score ${p_{t}}$. If the classifier is trained sufficiently, we can trust the model more with itself instead of the annotations. In this manner, the model would actually learn as if the anchor is positively labeled.

% Below is an example of how to insert images. Delete the ``\vspace'' line,
% uncomment the preceding line ``\centerline...'' and replace ``imageX.ps''
% with a suitable PostScript file name.
% -------------------------------------------------------------------------

% \begin{figure}[htb]

% \begin{minipage}[b]{1.0\linewidth}
%   \centering
%   \centerline{\includegraphics[width=8.5cm]{image1}}
% %  \vspace{2.0cm}
%   \centerline{(a) Result 1}\medskip
% \end{minipage}
% %
% \begin{minipage}[b]{.48\linewidth}
%   \centering
%   \centerline{\includegraphics[width=4.0cm]{image3}}
% %  \vspace{1.5cm}
%   \centerline{(b) Results 3}\medskip
% \end{minipage}
% \hfill
% \begin{minipage}[b]{0.48\linewidth}
%   \centering
%   \centerline{\includegraphics[width=4.0cm]{image4}}
% %  \vspace{1.5cm}
%   \centerline{(c) Result 4}\medskip
% \end{minipage}
% %
% \caption{Example of placing a figure with experimental results.}
% \label{fig:res}
% %
% \end{figure}

%\vfill
%\pagebreak

\section{Experiments and Analysis}
\label{sec:exps}

\subsection{Experimental Setting and Baselines}
\label{ssec:baseline}

We start our experiments with a baseline for benchmarks. First, we combine the PASCAL VOC 2007 and 2012 dataset (train + val) for training while using PASCAL VOC 2007 test set for inference. All objects marked as ``difficult'' in the original dataset are added for both training and testing. Then we erase some of the annotations in the training set to generate the following datasets as shown in Fig.~\ref{fig:dataset}: 

\vskip 0.05in
   \noindent(i) \textit{\textbf{normal}}: All annotations are preserved for training.

   \noindent(ii) \textit{\textbf{easy}}: One annotation is randomly removed in an image, providing multiple annotations are present.   

   \noindent(iii) \textit{\textbf{hard}}: Half of the annotations are randomly removed in an image, providing multiple annotations are present. 

   \noindent(iv) \textit{\textbf{extreme}}: Only one annotation is kept inside an image. 
\vskip 0.05in

We argue that at least one annotation should be preserved in each image to prevent unstable gradients while training with a small batch size. So, we start with four baseline detectors following the RetinaNet pipeline with ResNet-101 as backbone \cite{lin2017focal}. Specifically, each detector is trained on one custom dataset respectively for 12 epochs with a batch size of 16 on 8 Nvidia Titan X Pascal GPUs. The performance of the RetinaNet drops significantly as the amount of missing annotations increases (see Tab. \ref{tab:baseline}). This observation is anticipated since single stage anchor-based detectors suffer the most from incomplete dataset as every anchor contributes to the gradient propagation. 

\begin{table}[h]
\vspace{-0.02in}
\centering
\resizebox{.48\textwidth}{!}{
\begin{tabular}{l|cccc}
\Xhline{1pt}
\hline 
    \bf Dataset    & \bf Annotations Dropped    & \bf Baseline mAP 50 & \bf Baseline mAP 75             \\    \hline
    \bf \textit{normal}     & 0\%                        & 0.768               & 0.530                  \\
    \bf \textit{easy}       & 20.60\%                    & 0.716               & 0.463                  \\
    \bf \textit{hard}       & 39.00\%                    & 0.674               & 0.441                  \\
    \bf $\textit{extreme}$    & 64.95\%                    & 0.598               & 0.383                  \\    \Xhline{1pt}

\end{tabular}}
\vspace{-0.1in}
\caption{Baseline Detectors trained on four custom datasets with {\em Focal Loss}. All baseline detectors are trained on the \textbf{\textit{extreme}} dataset using the same parameters.}
\vspace{-2.1em}
\label{tab:baseline}
\end{table}

\subsection{Isolation of confusing anchors}
\label{ssec:confusing anchors}
To examine whether the adverse effect originates from confusing anchors, a detector is trained such that it only propagates loss from the positive and negative anchors. This behaves effectly as hard effective mining. We pick IoU of 0.1 as the threshold to isolate the confusion region according to \cite{wu2018soft}. To our surprise, the result is similar to the baseline performance. We then increase the loss contribution of the confusion anchors to half of the normal anchors with the proposed BRL and observe great improvement over the baseline. Because the confusion anchors count up towards a great portion of the total amount of anchors, simply removing its effect completely is less ideal. To locate the optimal IoU threshold, we conduct experiments under the IoUs of 0.2 and 0.05 and found no significant impact, as shown in Tab. \ref{tab:IoU}. This suggests that {BRL} can be fairly robust over both densely packed and sparsely arranged objects in the images. For the rest of the experiments, we use IoU 0.1 as the confusion IoU threshold. 
\begin{table}[h]
\centering
\resizebox{.45\textwidth}{!}{%
\begin{tabular}{l|c|c|cc}
\Xhline{1pt}
\hline
\multicolumn{5}{c}{\bf Performance on PASCAL VOC 2007 Test}           \\ \hline
\bf Pos IoU & \bf Neg IoU      & \bf Confusion IoU & \bf mAP$_{\bf 50}$ & \bf mAP$_{\bf 75}$  \\ \hline
$>$0.5     &  0.2 $\sim$ 0.4  &  $<$ 0.2         & 0.627       & 0.407        \\ 
$>$0.5     &  0.1 $\sim$ 0.4  &  $<$ 0.1         & 0.629       & \bf 0.409        \\
$>$0.5     &  0.05 $\sim$ 0.4 &  $<$ 0.05        & \bf 0.633       & 0.404        \\ \Xhline{1pt} \hline
\end{tabular}}
\vspace{-0.1in}
\caption{Locating the confusion anchors with IoU threshold of 0.05, 0.1, and 0.2. All experiments are conducted under the same settings except for the confusion IoU.}
\vspace{-1.em}
\label{tab:IoU}
\end{table}

\subsection{Background Recalibration Loss}
To investigate the proposed loss function, we perform the following ablation studies. We are interested in two factors: where to set the recalibration threshold $t$ and how to recalibrate? In brief, we assign threshold $t$ to several values, denoted as \textbf{$BRL_t$}. In the case of $t=0$, the loss is the equivalence of \textit{focal loss}. Additionally, the effectiveness of recalibration is studied by switching off the hard-negative part by setting the loss to $0$, marked as $BRL^-$. 

\begin{table}[h]
        \begin{minipage}{.21\textwidth}
        \centering
            \resizebox{1.0\textwidth}{!}{
            \begin{tabular}{l|ccc}
            \Xhline{1pt}
            \hline
            \bf experiment      & \bf mAP$_{\bf 50}$     & \bf mAP$_{\bf 75}$       \\ \hline
            \bf BRL$_{t=0}$         & 0.621              & 0.393                    \\
            \bf BRL$_{t=0.1}$         & 0.617               & 0.397                 \\
            \bf BRL$_{t=0.25}$         & 0.620              & 0.399                 \\
            \bf BRL$_{t=0.50}$        & \bf 0.629             & \bf 0.409           \\ \hline
            \bf BRL$_{t=0.50}^{-}$           & 0.622                      & 0.404            \\
            \Xhline{1pt}
            \hline
            \end{tabular}} 
    \end{minipage}
    \begin{minipage}{0.28\textwidth}
    
        \resizebox{0.9\textwidth}{!}{
        \begin{tabular}{l|cc}
        \Xhline{1pt}
        \multicolumn{3}{c}{\bf Performance on PASCAL VOC 2007 Test}     \\
        \Xhline{1pt}
        \bf loss weight & \bf mAP$_{\bf 50}$ & \bf mAP$_{\bf 75}$       \\ \hline
        1.0  & 0.608  & 0.398  \\
        0.5  & 0.629  & 0.409  \\
        0.25 & 0.651  & 0.419  \\
        0.1  & \bf 0.661  & \bf 0.431  \\
        0.01 & 0.631  & 0.417  \\
        \Xhline{1pt}
        \hline
        \end{tabular}}
    \end{minipage}
    \caption{Ablation Experiments on recalibration threshold $t$ and the scaling factor of the proposed {BRL}. In the recalibration threshold experiments, we trained all the detectors with scaling factor ${\alpha_{t}}$ of 0.5. In the ablation study of the scaling factor, we pick $t=0.5$ as it shows the best performance on \textbf{\textit{extreme}} dataset.}
    \vspace{-0.5em}
    \label{tab:thresh}
\end{table}

\vskip 0.05in
Among all the experiments, we observe that the maximum improvement is at $t=0.5$ while lower values only lead to minor improvements. Higher thresholds, however, result in unstable gradients during early training stages and the model will not converge. With recalibration, the performance are consistently above the baseline. Easy to notice that the total loss contribution of the confusion anchors directly has a great impact on the performance. For example, even without adopting {BRL}, the result on the \textbf{\textit{extreme}} dataset is better than baseline when we downscale the loss contribution to $50\%$. 

\begin{figure}[htb]
      \centering
      \centerline{\includegraphics[width=6.9cm]{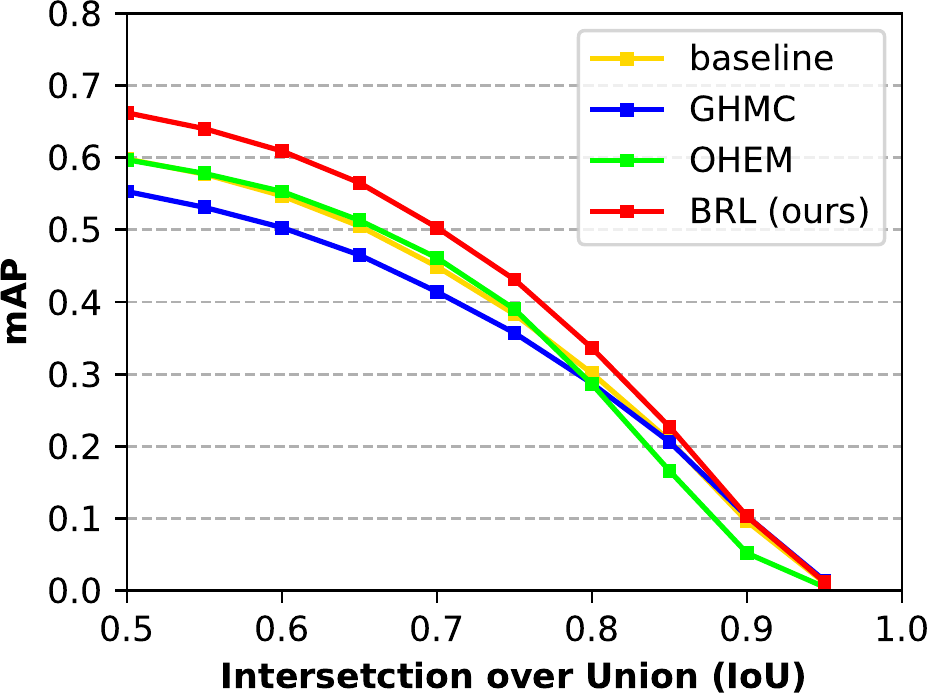}}
      \vspace{-.4em}
\caption{Consistant advantage across all IoU levels over the other sampling methods. All models trained on \textbf{\textit{extreme}} dataset.}
\vspace{-0.5em}
\label{fig:res}
\end{figure}

\vskip 0.05in
By lowering the loss contribution of the confusing anchors, the model suffers less from the missing labels. In Tab. \ref{tab:thresh}, we analyzing the {BRL} with a series of scaling factors. While using a scaling factor of 1.0, the model is not different from the baseline except for the recalibration scheme. We see about 1\% increase in performance on the \textbf{\textit{extreme}} dataset. This again, proves the effectiveness of the proposed recalibration. With a lower scaling factor of 0.1, we achieved the highest score, boosting mAP$_{50}$ and mAP$_{75}$ by 6.4\% and 4.8\% respectively.  
\vskip 0.05in
The improvements of mAP are consistant at IoU threshold above 0.5. The proposed method shows great advantages over common methods, such as GHMC \cite{li2019gradient} and FasterRCNN with OHEM \cite{shrivastavaCVPR16ohem}. The results are summarized in Tab. \ref{tab:stateofart} and Fig \ref{fig:res}. Furthermore, we study the performance on the modified COCO dataset where we randomly erase the annotations by 50\% per object category. Our model remains robust against missing labels and have a huge advantage over the baseline by 9.0\% in mAP$_{0.5:0.95}$ \cite{lin2014microsoft}. To see how the proposed loss performs on fully annotated datasets, we trained yet another detector with normal COCO dataset and found 0.7\% decrease in {mAP}. The scaling of gradients is possible to cause this minor performance drop indicating extended training time might be required.

\begin{table}
\centering
\resizebox{.48\textwidth}{!}{
\begin{tabular}{l|c|c|ccc}
\Xhline{1pt}
    ~~~~~~~~~~~~~~~~~\bf model & \bf train  & \bf test & \bf mAP$_{\bf 50}$ & \bf mAP$_{\bf 75}$ & \bf mAP$_{\bf 0.5:0.95}$ \\ \Xhline{1pt}
    \multirow{4}{6em}{RetinaNet+GHMC~\cite{li2019gradient}}   & \textbf{\textit{normal}} &  VOC 2007 Test & 0.754 & 0.532 & - \\
                & \textbf{\textit{easy}} &  VOC 2007 Test & 0.710 & 0.471 & -            \\
                & \textbf{\textit{hard}} &  VOC 2007 Test & 0.675 & 0.443 & -            \\
                & \textbf{\textit{extreme}} &  VOC 2007 Test & 0.553 & 0.357 & -    \\ \hline
    \multirow{4}{6em}{FasterRCNN+OHEM~\cite{shrivastavaCVPR16ohem}}  & \textbf{\textit{normal}} &  VOC2007 Test &\bf 0.778 & \bf 0.563 & - \\             
                & \textbf{\textit{easy}} &  VOC 2007 Test & \bf 0.751 & 0.486 & - \\
                & \textbf{\textit{hard}} &  VOC 2007 Test & 0.717 & 0.455 & - \\
                & \textbf{\textit{extreme}} &  VOC 2007 Test & 0.597 & 0.390 & - \\ \hline
    \multirow{4}{6em}{RetinaNet+BRL$_{t=0.50}$~(ours)}   & \textbf{\textit{normal}} &  VOC2007 Test & 0.753 & 0.532 & - \\
                & \textbf{\textit{easy}} &  VOC 2007 Test & 0.735 & \bf 0.501 & -            \\
                & \textbf{\textit{hard}} &  VOC 2007 Test & \bf 0.717 & \bf 0.479 & -            \\
                & \textbf{\textit{extreme}} &  VOC 2007 Test &\bf 0.662 &\bf 0.431 & -    \\ \Xhline{1pt}
    
    RetinaNet+FL &\bf coco$_{\bf 50\%}$ &  coco 2017 minival & 0.406  & 0.305 & 0.237 \\

    RetinaNet+BRL$_{t=0.50}$ (ours)   &\bf coco$_{\bf 50\%}$ &  coco 2017 minival & \bf 0.508 & \bf 0.353 & \bf 0.327  \\ \Xhline{1pt}
    
\end{tabular}}
\vspace{-0.6em}
\caption{Performance comparisons on PASCAL VOC 2007 and COCO 2017 minival. Our method achieves excellent performance and surpasses two-stage detectors FasterRCNN with OHEM on both difficult datasets. It is also noticeable the {BRL} does not significantly hurt the performance while training with fully annotated dataset.}
\vspace{-0.5em}
\label{tab:stateofart}
\end{table}    

\vspace{-1.em}
\section{Conclusion}
\label{sec:conclusion}
 \vspace{-.4em}
We have presented {\em Background Recalibration Loss}, a novel loss function deign that is more fit for missing-labeled object detection scenario, and improves the detection performance vastly. Extensive experiments on PASCAL VOC and MS COCO demonstrate the effectiveness of our proposed method. Since  training with missing-labeled objects usually uses limited instances data, our future work will focus on adopting GAN-based image generation and data augmentation method like MUNIT~\cite{huang2018multimodal}, INIT~\cite{shen2019towards}, etc. to  generate pseudo labels of objects and enlarge the diversity of training samples, in order to obtain better detection performance when learning under sparse annotations.

% \vfill\pagebreak

\bibliographystyle{IEEEbib}
\bibliography{strings,refs}

\end{document}